\pdfoutput=1

\documentclass[11pt]{article}

\usepackage[final]{acl}

\usepackage{longtable}
\usepackage{geometry}
\usepackage{graphicx}
\usepackage{adjustbox}
\usepackage{subfigure}
\usepackage{subcaption}
\usepackage{array,booktabs,ragged2e}
\usepackage{multirow}
\usepackage{multicol}
\usepackage{times}
\usepackage{latexsym}
\usepackage{arydshln}
\usepackage{pifont}
\usepackage{float}
\usepackage{bm}
\usepackage{microtype}
\usepackage{inconsolata}
\usepackage{xcolor}
\usepackage{xspace}

\usepackage{placeins}
\usepackage{afterpage}
\usepackage{amsfonts}
\usepackage{amsmath}

\usepackage[T1]{fontenc}
\usepackage[utf8]{inputenc}

\usepackage{booktabs}       
\usepackage{adjustbox}
\usepackage{multirow}
\usepackage{siunitx}

\usepackage{amsmath}
\usepackage{amssymb}
\usepackage{mathtools}
\usepackage{amsthm}
\usepackage{mathabx}
\usepackage{array}

\usepackage{enumitem}
\usepackage{subfigure}
\setlist[enumerate]{nosep}
\usepackage{soul}
\usepackage{caption}
\usepackage{xcolor}
\usepackage{xspace}
\usepackage{arydshln}
\usepackage{pifont}

\usepackage{bm}

\definecolor{Ind}{HTML}{F17D7D}
\definecolor{US}{HTML}{3C6C76}
\definecolor{Trans}{HTML}{9A70ED}
\definecolor{Gen}{HTML}{458657}
\definecolor{Red}{HTML}{AC2037}
\definecolor{Gray}{gray}{0.8}
\definecolor{filler}{HTML}{9F080E}
\definecolor{rephrase}{HTML}{F7AB1E}
\definecolor{mask}{HTML}{9F080E}
\definecolor{prune}{HTML}{9F080E}

\newcommand{\IndEng}{en-IN\xspace}
\newcommand{\USEng}{en-US\xspace}
\newcommand{\Trans}{en-TR\xspace}
\newcommand{\Multi}{en-MV\xspace}

\newcommand{\gptf}{\textsc{gpt-4}\xspace}
\newcommand{\gptt}{\textsc{gpt-3.5}\xspace}
\newcommand{\llama}{\textsc{llama-3}\xspace}

\newcommand{\md}{MD3\xspace}
\newcommand{\mmd}{M-MD3\xspace}

\newcommand{\na}{\texttt{\textbf{-}}\xspace}

\newcommand{\ind}{%
  \begingroup\normalfont
  \includegraphics[height=11px]{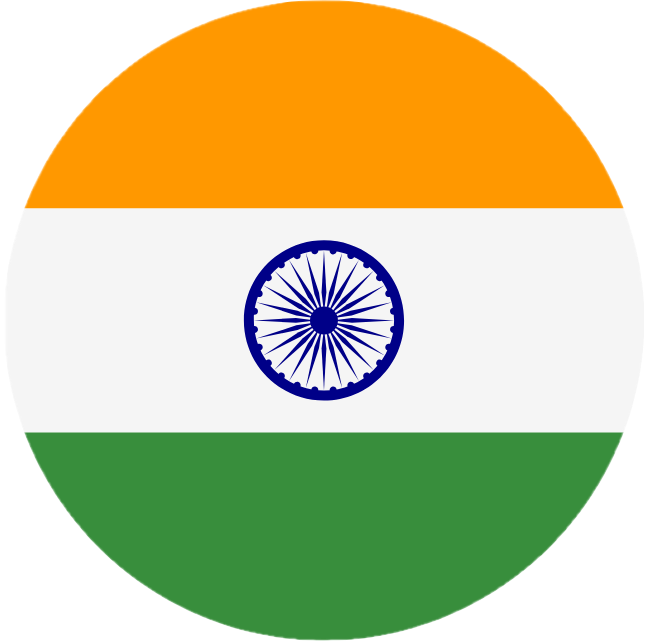}%
  \endgroup
}
\newcommand{\us}{%
  \begingroup\normalfont
  \includegraphics[height=11px]{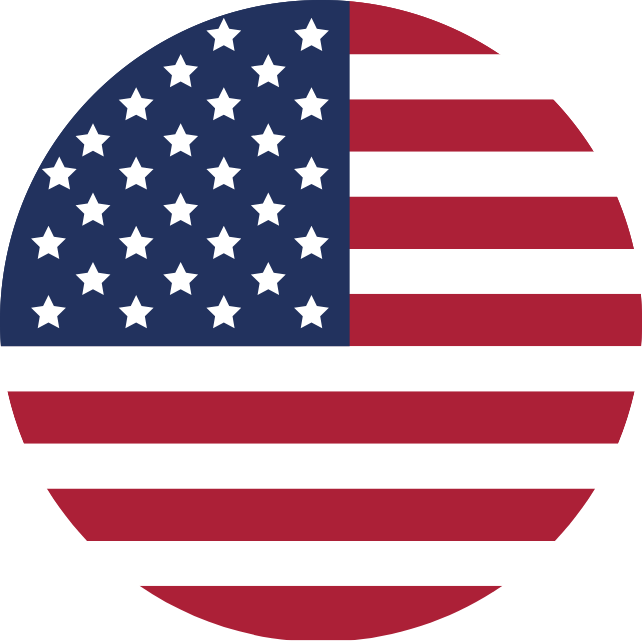}%
  \endgroup
}
\newcommand{\desc}{%
  \begingroup\normalfont
  \includegraphics[height=11px]{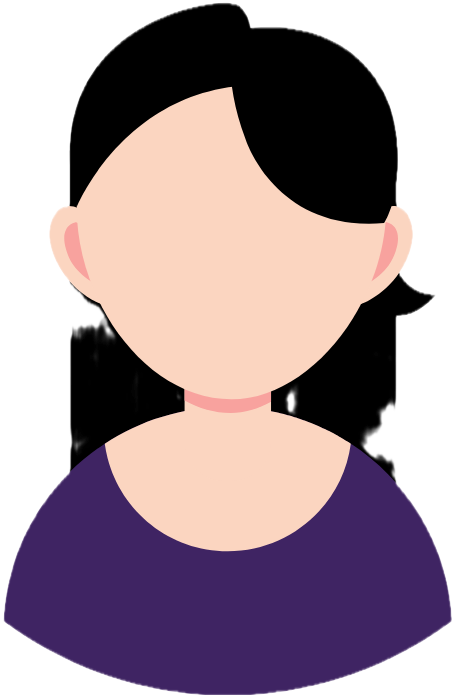}%
  \endgroup
}
\newcommand{\gues}{%
  \begingroup\normalfont
  \includegraphics[height=11px]{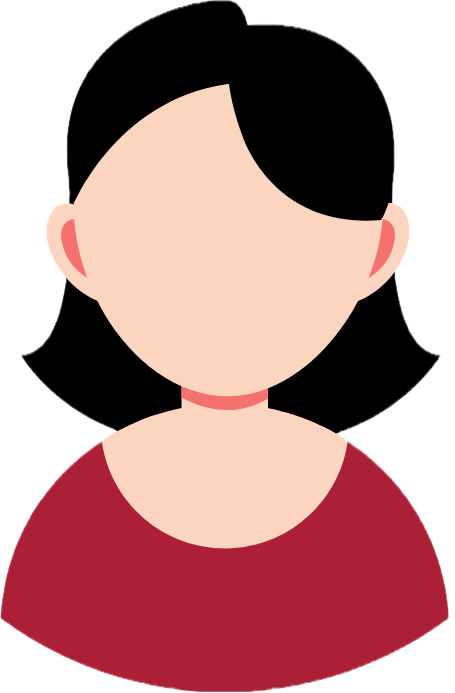}%
  \endgroup
}

\title{Evaluating Dialect Robustness of Language Models via Conversation Understanding}
\author{
    Dipankar Srirag\textsuperscript{\rm 1}\quad Nihar Sahoo\textsuperscript{\rm 2}\quad Aditya Joshi\textsuperscript{\rm 1}\\
    \textsuperscript{\rm 1}University of New South Wales, Sydney, Australia\\
    \textsuperscript{\rm 2}Indian Institute of Technology Bombay, India\\
    \texttt{\{d.srirag, aditya.joshi\}@unsw.edu.au}\quad\texttt{nihar@cse.iitb.ac.in}
}

\begin{document}
\maketitle
\begin{abstract}
With an evergrowing number of LLMs reporting superlative performance for English, their ability to perform equitably for different dialects of English (\textit{i.e.}, dialect robustness) needs to be ascertained. Specifically, we use English language (US English or Indian English) conversations between humans who play the word-guessing game of `taboo`. We formulate two evaluative tasks: target word prediction (TWP) (\textit{i.e.}, predict the masked target word in a conversation) and target word selection (TWS) (\textit{i.e.}, select the most likely masked target word in a conversation, from among a set of candidate words). Extending \md, an existing dialectic dataset of taboo-playing conversations, we introduce \mmd, a target-word-masked version of \md with the \USEng and \IndEng subsets. We create two subsets: \Multi (where \USEng is transformed to include dialectal information) and \Trans (where dialectal information is removed from \IndEng). We evaluate three multilingual LLMs--one open-source (Llama3) and two closed-source (GPT-4/3.5). LLMs perform significantly better for US English than Indian English for both TWP and TWS tasks, for all settings, exhibiting marginalisation against the Indian dialect of English. While GPT-based models perform the best, the comparatively smaller models work more equitably after fine-tuning. Our evaluation methodology exhibits a novel and reproducible way to examine attributes of language models using pre-existing dialogue datasets with language varieties. Dialect being an artifact of one's culture, this paper demonstrates the gap in the performance of multilingual LLMs for communities that do not use a mainstream dialect.
\end{abstract}
\section{Introduction}\label{sec:intro}
\begin{figure}[t!]
    \begin{adjustbox}{width=1\linewidth,height=0.83\linewidth, center}
        \includegraphics{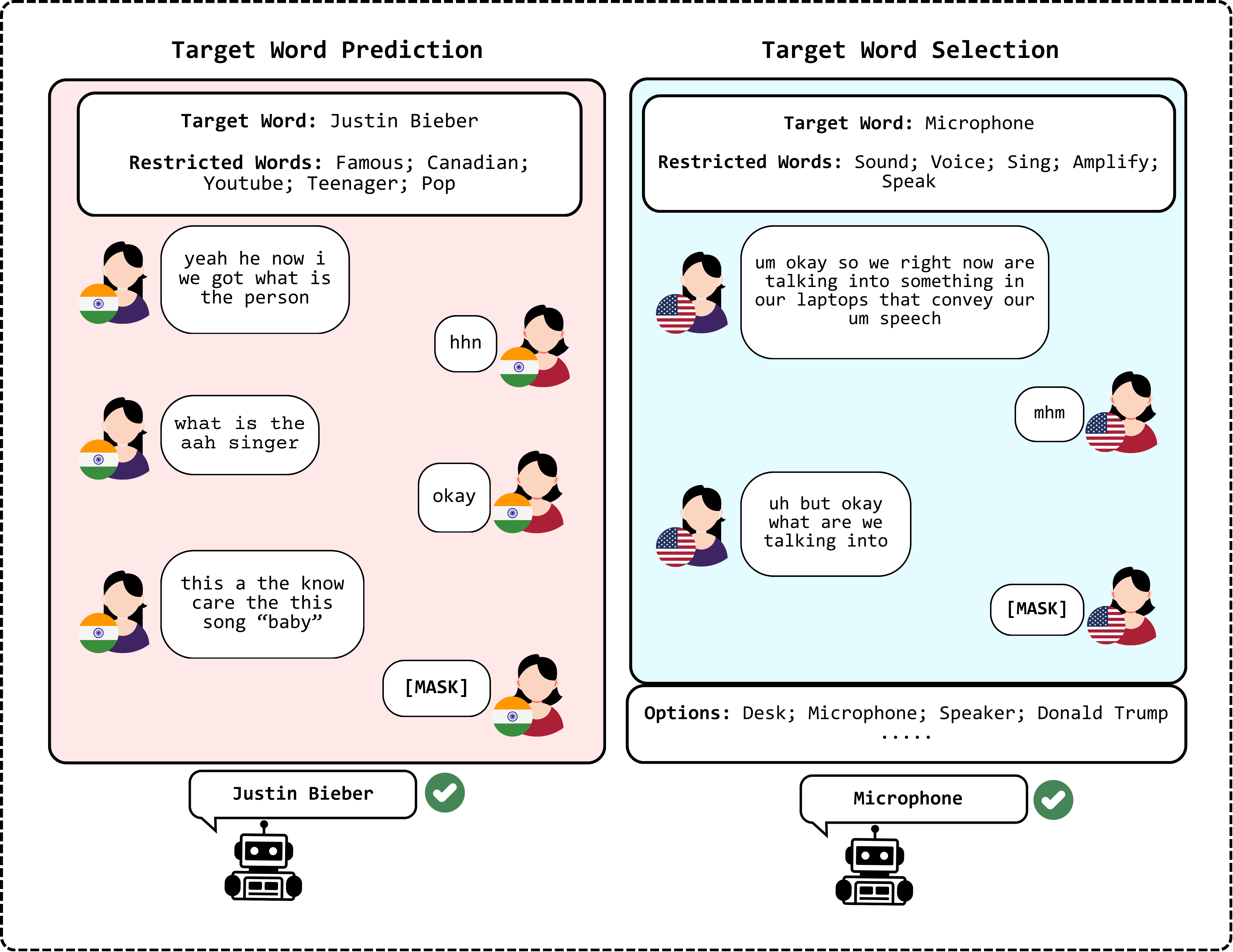}
    \end{adjustbox}
    \caption{Illustration of the two tasks: Target word prediction (TWP) and Target word selection (TWS).~\desc  and~\gues are the describer and the guesser respectively in a word-guessing game of taboo.~\ind and ~\us refer to Indian English and US English respectively.}
    \label{fig:scenario}
\end{figure}
Large language models (LLMs)\footnote{We use `language models' and `large language models/LLMs' interchangeably in this paper.} based on Transformers~\cite{NIPS2017_3f5ee243} are the state-of-the-art in natural language processing (NLP), often reporting superlative performance on several NLP tasks~\cite{zhao2023survey}. These models predominantly use English language data in their pre-training corpus. However, being a widely spoken language, English takes multiple forms in different parts of the world. These forms, called dialects or national varieties of English, collectively constitute the World Englishes~\cite{bolton2012world}. While research papers introducing LLMs report performance on English language datasets, recent works highlight the performance gap between US English and other dialects of English for several natural language processing tasks~\cite{joshi2024natural}. 

Our paper examines cultural considerations of evaluating LLMs through the prism of dialect robustness via conversation understanding. The choice of conversation understanding as a domain for evaluation emerges from the fact that dialectal features are most visible in free-flowing conversations~\cite{negro_vietti_2006}. Therefore, we investigate the research question:

\begin{quote}
    ``\textit{In comparison with US English, how effectively can LLMs understand conversations between speakers of other national varieties of English?}''
\end{quote}

To address the research question, we use a pre-existing dataset-- \textbf{\md}~\cite{eisenstein2023md3} that consists of manually transcribed dialogues between pairs of human participants where each pair speaks either Indian English or US English. The participants engage in a focused conversation: they play the word-guessing game based on the game of `Taboo'~\cite{taboo-wiki}. In the game, a describer must get a guesser to identify a \textbf{target word} but must not use a set of related words known as \textbf{restricted words} while describing the target word. 
Using this dataset of dialectal dialogues, we introduce two tasks to evaluate the dialect-robustness of LLMs to understand conversations. They are: (a) Given an input conversation with the target word masked, can the LLM \textit{predict} the target word? (referred to as \textbf{target word prediction}) (b) Given an input conversation with the target word masked along with a set of candidate target words, can the LLM \textit{select} the correct target word? (referred to as \textbf{target word selection}). Our approach of masking the target word is similar to~\citet{dey-desarkar-2023-dial}, who show that masked word prediction may correlate with automatic dialogue evaluation metrics.
 Figure~\ref{fig:scenario} shows an example of the two tasks, where the language model predicts `\textit{Justin Bieber}' for target word prediction, and selects `\textit{microphone}' among the set of options for target word selection\footnote{We run experiments on both the tasks for both US and Indian English conversations. While the examples show expected output, the LLM may or may not produce the same in the case of our experiments. That is the crux of the evaluation.}.
For the two tasks, we extend \md to create a target-word-\textbf{M}asked \textbf{M}ulti-\textbf{D}ialect \textbf{D}ataset of \textbf{D}ialogues (\textbf{\mmd})\footnote{\mmd dataset and the related code will be made publicly available at ~\url{https://github.com/dipankarsrirag/eval-dialect-robust}.}. \mmd consists of (a) conversations between Indian English speakers (\IndEng), and conversations between US English speakers (\USEng), (b) \USEng conversations transformed into \IndEng using rule-based perturbations (\Multi), (c) \IndEng with dialectal information removed (\Trans). We evaluate the performance of three SOTA large language models (LLMs), one open-source and two closed-source, employing zero-shot prompting on both pre-trained and fine-tuned models (where available).
Our evaluation methodology derives from past work that evaluates LLMs by providing a set of task-specific examples~\cite{wang-etal-2023-self-instruct}.
Of particular relevance is the work by ~\citet{chalamalasetti-etal-2023-clembench}, who generate word game conversations using LLMs and evaluate their ability to predict the target word.
The contributions of our work are:
\begin{itemize}
 \item We create \mmd, an extension of \md, that deals with two novel evaluative tasks for dialect robustness: target word prediction and target word selection.
 \item Our evaluation demonstrates a degraded performance in the case of Indian English as compared to US English for all models, supporting existing social disparities between US and Indian culture in the LLM representations~\cite{10.1145/3677525.3678666}.

 \item A comprehensive error analysis to identify specific conditions under which fine-tuning enhances the model’s performance on Indian English conversations.
\end{itemize}
Since several LLMs have been deployed as publicly available dialogue agents\footnote{ChatGPT~\url{https://chat.openai.com/}; Accessed on 9th April 2024.}, it is imperative that they can understand the conversations of users belonging to diverse English-speaking subgroups. In the case of our paper, this refers to dialectal variations, considering them as a proxy to culture. The rest of the paper is organized as follows. Section~\ref{sec:method} introduces our evaluation methodology. The experiment setup and results are in Sections~\ref{sec:setup} and ~\ref{sec:results} respectively.

\section{Methodology}\label{sec:method}
\begin{figure*}[t]
    \centering
    \includegraphics[width=1\linewidth]{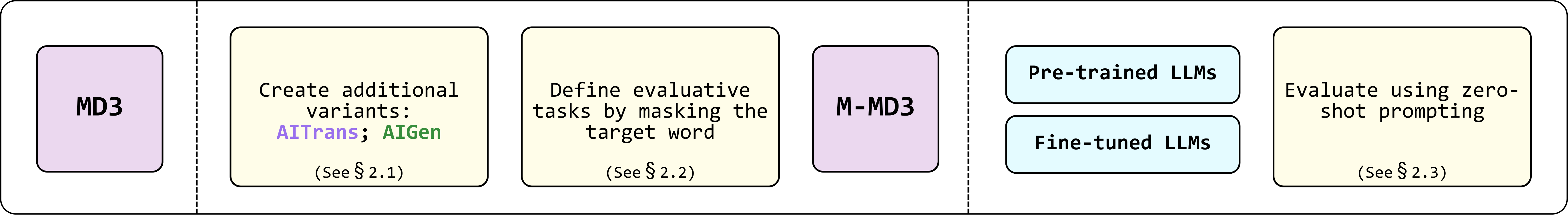}
    \caption{Steps for evaluation of dialect robustness.}
    \label{fig:method}
\end{figure*}

We present our method step-by-step, with a detailed overview of our evaluation methodology described in Figure~\ref{fig:method}. We select two subsets available in \md: \IndEng and \USEng, and filter out the conversations where the guesser could not identify the target word. We extend \md to include two additional sets of conversations---\Multi and \Trans, and mask the target words in all four subsets to create \mmd. We ensure that the mask token always appears at the end of the conversation, warranting the use of auto-regressive models. This is done by pruning the conversation to the turn where the guesser utters the target word\footnote{Details on the masking method with examples are provided in Appendix~\ref{sec:masking}.}. 

Transforming text in \USEng to other dialectal English text has been explored for low-resource settings~\cite{held-etal-2023-tada, xiao-etal-2023-task, liu-etal-2023-dada}. To evaluate the efficacy of synthetically transformed dialogues, we extend the dataset of dialectal dialogues to include two additional sets of conversations-- \Multi and \Trans.
\paragraph{\Multi} We use Multi-VALUE~\cite{ziems2023multi} to transform \USEng conversations into \IndEng conversations. We call this set of conversations created by rule-based transformations \Multi.
\paragraph{\Trans} We prompt GPT-4 Turbo Preview(\gptf;~\citeauthor{openai2024gpt4}~\citeyear{openai2024gpt4}) to remove dialectal information from \IndEng. The resultant set of conversations is known as \Trans. The prompt\footnote{The forms are experimentally determined using a few test examples.} used to generate such conversations is given below:
\begin{quote}
    ``\textit{Normalise the conversation. Remove all exaggerations and dialectal information. Return a neutral response.}''
\end{quote}
\subsection{Extending \md} \label{sec:dataset}

The use of \gptf to transform \IndEng conversations sometimes leads to the generation of conversation summaries rather than transformed conversations\footnote{More details with examples are discussed in Appendix~\ref{sec:issues}.}. Due to the varying lengths of speaker turns, transforming \USEng conversations using Multi-VALUE occasionally fails to output a result.  Such failed transformations are excluded from both the subsets of transformed (\Multi, \Trans) conversations, leading to fewer conversations in \Multi and \Trans as compared to \USEng and \IndEng, respectively, as shown in Tables \ref{tab:add_stats} and \ref{tab:stats}.
\begin{figure}[h!]
    \begin{adjustbox}{width=1\linewidth, center}
        \includegraphics{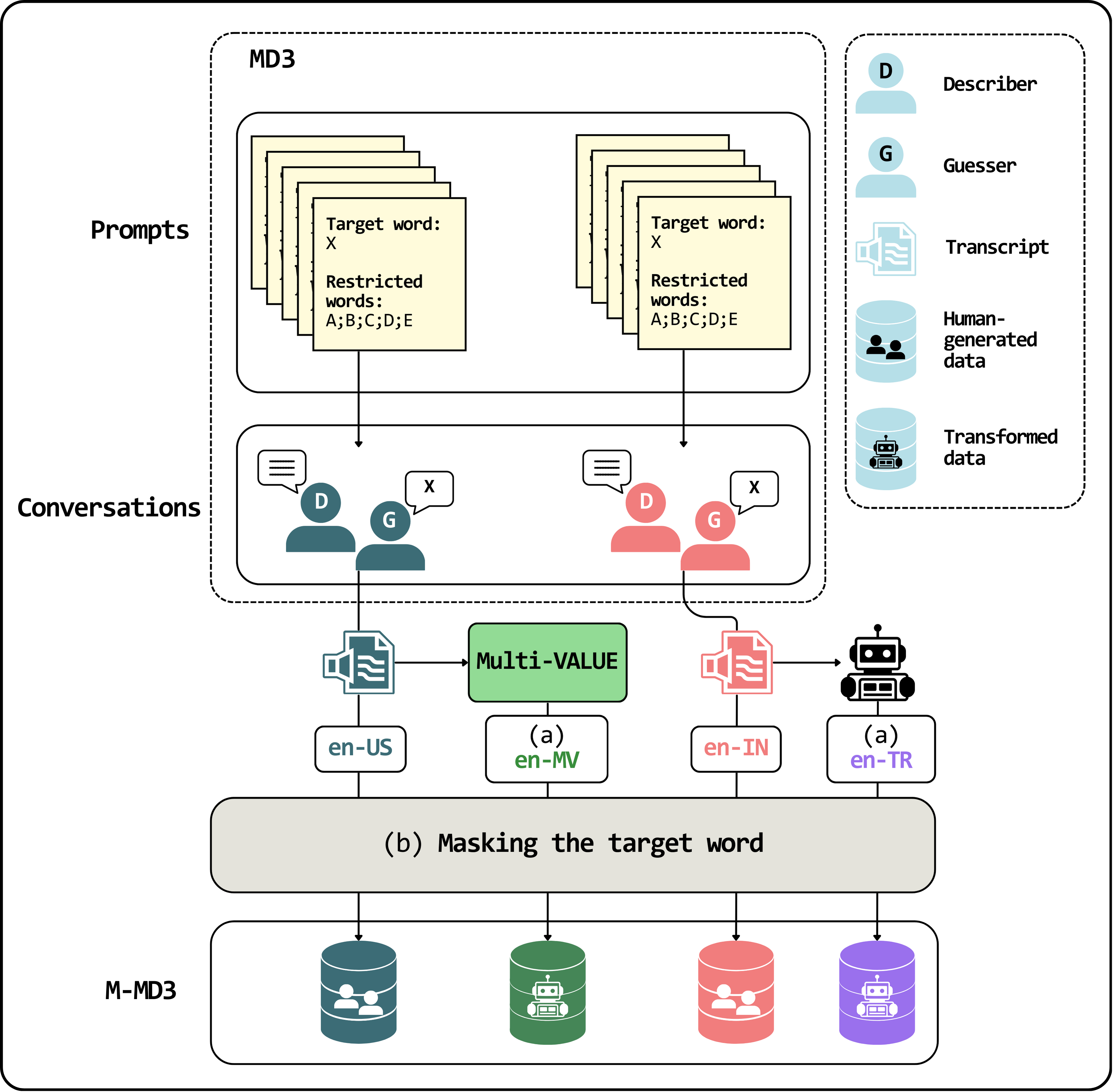}
    \end{adjustbox}
    \caption{\mmd as an extension of \md: (a) Creation of \Multi and \Trans, and (b) Creation of target-word-masked conversations.}
    \label{fig:dataset}
\end{figure} 

\subsection{Analysis}
\label{sec:analysis}
Table \ref{tab:add_stats} reports some of the constructional statistics of \mmd. For each subset, it reports the average number of dialogue turns per conversation, the average word count for the dialogues uttered by both the describer and the guesser, and the number of conversations with single-word versus multiple-word reference target words. The target words `\textit{microphone}' and `\textit{Justin Bieber}' in Figure~\ref{sec:intro} are examples of single-word and multiple-word reference target words, respectively.

We notice a higher number of average turns and words spoken in \IndEng conversations compared to \USEng conversations. This is due to the \USEng speakers being more familiar with the target word compared to \IndEng speakers, leading to shorter gameplay time~\cite{eisenstein2023md3}. The trend is also carried over to the transformed conversations in \Multi (derived from \USEng) and \Trans (derived from \IndEng).
\begin{table}[]
     \begin{adjustbox}{width=1.0\columnwidth,center}
     \scriptsize{
         \begin{tabular}{ccccc}
         \toprule
         Subset & Avg. turns & Avg. words & Single & Multiple\\\midrule[\heavyrulewidth]
         \USEng & 4.1 & 42.1& 308 & 106\\
         \IndEng & 6.8 & 57.4 & 153 & 59\\
         \Multi & 4.9 & 35.2 & 245 & 87\\
         \Trans & 6.3 & 42.7 & 121 & 50\\
         \bottomrule
         \end{tabular}
         }
     \end{adjustbox}
     \caption{Constructional Statistics of \mmd. \textit{Single} and \textit{Multiple} refer to the number of conversations with single-word and multiple-word reference targets, respectively.}
     \label{tab:add_stats}
 \end{table}
\subsection{Task Definition} \label{sec:task}
 As shown in Figure~\ref{fig:dataset}, we mask the target word in the conversations from all four subsets. The target word occurs in the last dialogue turn of the conversation, which is spoken by the guesser\footnote{This always holds because of the way we process the conversations.}. As a result, we formulate two tasks where the expected output is to fill the correct word at the masked position:
\begin{itemize}
    \item \textbf{Target Word Prediction (TWP)}: Given a conversation with the target word masked, predict the word.
    \item \textbf{Target Word Selection (TWS)}: Given a conversation with the target word masked and a set of candidate target words, select one among the candidate set.
\end{itemize}

In the case of TWP, the LLM may generate any word within its learned vocabulary, with the expected output being the reference target word. In the case of TWS, we provide the LLM with a masked conversation and a set of all target words in the \mmd corpus. The LLM must then select the most likely target word. 

We then use prompting on three LLMs to perform both tasks (TWP and TWS). As LLMs, we choose models that have been optimised to follow natural language instructions. In our case, the instruction is to either predict the masked target word or select a word from candidate words. Specifically, we use one open-source model, namely, Llama 3 70B Chat (\llama;~\citeauthor{dubey2024llama3herdmodels}~\citeyear{dubey2024llama3herdmodels}), and two closed-source models, namely, \gptf and GPT-3.5 Turbo 0125 (\gptt;~\citeauthor{ouyang2022training}~\citeyear{ouyang2022training}).
\section{Experiment Setup}\label{sec:setup}

We report the performance on pre-trained and fine-tuned versions of multilingual LLMs using zero-shot prompting. Fine-tuning is always done `\textit{in-dialect}' in our case, although there is no reason to believe that cross-dialect fine-tuning is not possible.
 
\subsection{Model Parameters} 
Experiments on \gptf and \gptt are conducted using OpenAI's API\footnote{OpenAI API~\url{https://platform.openai.com/docs/api-reference}; Accessed on 18th April 2024.}. \gptt is fine-tuned for 5 epochs, separately for every subset.
We select top\_p as 0.2 to restrict variability in output generation. 

\llama is fine-tuned for 20 epochs, with a batch size of 16, Paged 8-bit AdamW~\cite{dettmers20228bit} as the optimiser and a learning rate of 2e-4. We use QLoRA adaptors, targeting all linear layers, as recommended by~\citet{dettmers2023qlora}. All experiments on \llama were performed using two A100 GPUs.

\subsection{Metrics}\label{sec:metrics}
We report our results on two metrics: \textit{accuracy} and \textit{similarity}. \textit{Accuracy} is the proportion of conversations where the LLM generated the correct target word. This is a strict metric in that it requires the LLM to generate an exact match to the reference target word. In the case of TWP, the LLM will choose from all the words within its vocabulary, while in the case of TWS, the LLM will choose from the set of candidate target words. Therefore, it is trivial that the accuracy for TWS is expected to be higher than that for TWP. Accuracy metric penalizes models even if the generated target word partially matches with the reference target word in case of multi-word reference target as described in Section \ref{sec:analysis}. As \textit{similarity}, we report the cosine similarity between the Sentence-BERT embeddings~\cite{reimers2019sentencebert} of the reference target word and the generated target word. This allows for similar but inexact words generated by the LLM to be acceptable to the similarity score.

\subsection{Experiments}

We perform experiments on both the tasks (TWP and TWS) using all models(\{pre-trained and fine-tuned\} $\times$ \{\gptf, \gptt, \llama\}). All results are reported only on the test split of each subset of conversations. All fine-tuned models are fine-tuned on the training and validation set using instruction fine-tuning. \gptf could not be fine-tuned because doing so is restricted by OpenAI at the time of writing this paper. The statistics of \textbf{Train}, \textbf{Valid}, and \textbf{Test} splits of each subset of \mmd are reported in Table \ref{tab:stats}.

\begin{table}[h]
     \begin{adjustbox}{width=0.7\columnwidth,center}
     \scriptsize{
         \begin{tabular}{cccc}
         \toprule
         Subset & Train & Valid & Test\\\midrule[\heavyrulewidth]
         \USEng& 62  & 41 & 311\\
         \IndEng& 31  & 21 & 160\\
         \Multi& 49  & 33 & 250\\
         \Trans & 23 & 17 & 131\\
         \bottomrule
         \end{tabular}
         }
     \end{adjustbox}
     \caption{Statistics of \mmd.}
     \label{tab:stats}
 \end{table}

\section{Results}\label{sec:results}
In this section, we compare the performance of three LLMs both quantitatively and qualitatively. Note that the same test split is used to evaluate both pre-trained and fine-tuned versions, ensuring that the results are comparable.

\subsection{Quantitative Results}
Table~\ref{tab:results} shows the results of our experiments on each task specified in Section~\ref{sec:task}. We analyse the results as follows.
\begin{table*}[t!]
    \begin{adjustbox}{width=1.0\linewidth,center}
        \setlength{\tabcolsep}{6pt}
        \setlength{\cmidrulekern}{0.25em}
        \begin{tabular}{cccccccccccccc}
        \toprule
        \multirow{3}{4.5em}{\centering{Model}}&\multirow{3}{4em}{\centering{Subset}}&\multicolumn{6}{c}{TWP} & \multicolumn{6}{c}{TWS}\\\cmidrule(lr){3-8}\cmidrule(lr){9-14}
        & &\multicolumn{3}{c}{Similarity}& \multicolumn{3}{c}{Accuracy} &\multicolumn{3}{c}{Similarity}& \multicolumn{3}{c}{Accuracy}\\\cmidrule(lr){3-5}\cmidrule(lr){6-8}\cmidrule(lr){9-11}\cmidrule(lr){12-14}
        & & PT & FT & $\Delta$ & PT & FT & $\Delta$ & PT & FT & $\Delta$ & PT & FT & $\Delta$\\\midrule[\heavyrulewidth]
        \multirow{5}{4.5em}{\centering\gptf} %
        & \USEng & \underline{\textbf{77.4}} & \na & \na & \underline{\textbf{67.8}} & \na& \na & \underline{\textbf{85.7}} & \na & \na & \underline{\textbf{78.8}} & \na & \na\\
        & \IndEng & \underline{63.0} & \na & \na & \underline{45.6} & \na & \na & \underline{79.0} & \na & \na & \underline{72.5} & \na& \na \\
        & \Multi & \underline{75.6} & \na & \na & \underline{60.0} & \na& \na & \underline{83.6} & \na & \na & \underline{74.4} & \na & \na \\
        & \Trans & \underline{62.8} & \na & \na & \underline{45.8} & \na& \na & \underline{83.4} & \na & \na & \underline{77.1} & \na& \na \\\cmidrule(lr){2-14}
        &$\delta$& -14.4 & \na & \na & -22.0 & \na & \na & -6.7 & \na & \na & -6.3 & \na & \na \\\midrule[\heavyrulewidth]
        \multirow{5}{4.5em}{\centering\gptt} %
        & \USEng & \textbf{66.3} & \textbf{72.2} & 5.9 & \textbf{52.7} & \textbf{59.1} & 6.4 & 66.4 & \textbf{80.8} & 14.4 & 50.8 & \textbf{71.3} & 20.5 \\
        & \IndEng & 53.2 & 59.1 & 5.9 & 34.4 & 40.0 & 5.6 & 61.9 & 70.7 & 8.8 & 47.5 & 60.6 & 13.1 \\
        & \Multi & 57.6 & 71.3 & 13.7 & 40.0 & 54.4 & 14.4 & 52.4 & 71.5 & 19.1 & 31.6 & 57.6 & 26.0 \\
        & \Trans & 59.4 & \underline{61.0} & 1.6 & 39.7 & 41.2 & 1.5 & \textbf{70.7} & 73.0 & 2.3 & \textbf{57.3} & 60.3 & 3.0\\\cmidrule(lr){2-14}
        &$\delta$& -13.1 & -13.1 & \na & -18.3 & -19.1 & \na & -4.5 & -10.1 & \na & -21.0 & -16.2 & \na \\\midrule[\heavyrulewidth]
        \multirow{5}{4.5em}{\centering\llama} %
        & \USEng & \textbf{70.8} & \underline{\textbf{78.0}} & 7.2 & \textbf{60.5} & \underline{\textbf{65.3}} & 4.8 & \textbf{78.0} & \underline{\textbf{81.8}} & 3.8 & \textbf{67.5} & \underline{\textbf{74.6}} & 7.1 \\
        & \IndEng & 59.8 & \underline{66.3} & 6.5  & 43.8 & \underline{54.4} & 10.6 & 68.8 & \underline{80.8} & 12.0 & 56.9 & \underline{74.4} & 17.5 \\
        & \Multi & 68.6 & \underline{73.8} & 5.2  & 54.0 & \underline{61.6} & 7.6 & 72.3 & \underline{77.6} & 5.3 & 58.8 & \underline{67.2} & 8.4 \\
        & \Trans & 60.7 & 57.5 & -3.2  & \underline{45.8} & \underline{42.7} & -3.1 & 70.8 & \underline{79.5} & 8.7 & 60.3 & \underline{72.5} & 12.2 \\\cmidrule(lr){2-14}
        &$\delta$& -11.0 & -11.7 & \na & -16.7 & -10.9 & \na & -9.2 & -1.8 & \na & -10.6 & -0.2 & \na \\
        \bottomrule
        \end{tabular}
    \end{adjustbox}
    \caption{Performance on the two tasks: TWP and TWS.  PT/FT: Pre-trained/Fine-tuned. {$\delta$} is the difference in performance between \IndEng and \USEng (\IndEng minus \USEng). {$\Delta$} is the difference in performance between FT and PT. The best performance by a model is represented with \textbf{bold} numbers. The best performance for a subset of conversations is represented with \underline{underlined} numbers.}
    \label{tab:results}
\end{table*}

\paragraph{\USEng versus \IndEng}
The focus of this paper is to evaluate dialect robustness by comparing the performance on \USEng and \IndEng. All LLMs perform consistently better on \USEng as compared to \IndEng for all configurations. For example, in the case of \llama and TWP, the similarity scores on the fine-tuned model are 78.0 for \USEng and 66.3 for \IndEng, with the drop in performance of 11.7. Even for all three models, \USEng outperforms \IndEng on zero-shot performance using the pre-trained model. From all results, it is clearly understood that, on average, the LLMs understand the US English dialect better than the Indian English dialect. Only considering the pre-training setting, \gptf outperforms other models for both \USEng and \IndEng. However, fine-tuning improves the performance of \llama on \IndEng, achieving better results on both tasks compared to GPT-based models. Interestingly, for \llama, the performance improvement after fine-tuning on \IndEng is greater compared to fine-tuning on \USEng (represented by $\Delta$). 

\paragraph{Impact of transforming conversations}
As discussed in section \ref{sec:dataset}, we introduced two synthetically transformed subsets, \Multi and \Trans, to assess the importance of dialectal features in LLMs' understanding of conversations. Table \ref{tab:results} shows that, on pre-trained models, \Trans conversations have better performance compared to original \IndEng conversations. This suggests that after removing the dialectal information from \IndEng, the resulting \Trans conversations are close to the distribution of the dialect that the LLM understands. This behaviour is better reflected in \gptt, potentially, because the LLM has a poor understanding of \IndEng as compared to the other two LLMs. Additionally, fine-tuning on \Trans conversations does not improve the task performances in comparison to that on \IndEng. This supports the hypothesis that the removal of dialectal information brings the resulting conversation closer to the dialectal distribution that LLMs understand than the original dialect. 

In the case of \Multi, the task performances are consistently lower compared to \USEng. For example,
in the case of \gptt and TWS, the similarity scores on
the fine-tuned model are 80.8 for \USEng and 71.5 for \Multi. This degraded performance shows that the rule-based transformation into \IndEng from \USEng reduced the understanding capacity of LLMs for the resulting conversations, further strengthening our hypothesis that {LLMs perform well for US English dialects compared to any other varieties}, similar to findings of \citet{ryan2024unintendedimpactsllmalignment}.

\paragraph{Shorter turns versus Longer turns} 
A trend appears between the performances of models on each subset of conversations and the constructional properties of these conversations discussed in Section~\ref{sec:analysis}. Models report their best performances on the subset with the smallest number of average turns in a conversation (\USEng), and report the worst performance on the subset with the highest number of average turns in a conversation (\IndEng).

\paragraph{TWP versus TWS}
We now compare the performances of TWP and TWS. As expected, the similarity and accuracy are higher in the case of TWS compared to TWP for all three models, with one exception: the pre-training performance of \gptt on \Multi, where TWP slightly outperforms TWS. Note that, for pre-trained \llama, the accuracy on \IndEng is 43.8 for TWP and 56.9 for TWS. Across all configurations, fine-tuning consistently improves the performance of both TWP and TWS. \gptf performs best (only for pre-trained models) for both TWP and TWS tasks for all subsets.

\paragraph{Model Comparison}
It can be easily observed from Table \ref{tab:results} that the \gptf outperforms the other two LLMs in the pre-training setting. Interestingly, for TWS, \gptf pre-training performances are better than fine-tuning performances of \gptt and \llama in most of the cases. Also, \gptf performs almost equally well for each subset of \mmd. This shows that \gptf and \llama are more inclusive for different dialectal variations of English in the pre-training and fine-tuning setting, respectively.

\paragraph{Pre-training versus Fine-tuning} 
Although the pre-training performances of \gptf are superlative, Table \ref{tab:results} shows that the fine-tuning also improves the performance of \gptt and \llama across both tasks and four subsets. Fine-tuning is more effective for \USEng than \IndEng in the case of \gptt, whereas \llama shows the opposite trend. 
For \gptt, the most improvement due to fine-tuning is seen when the models are fine-tuned on \Multi, while \llama shows the highest improvement when fine-tuned on \IndEng.

\subsection{Error Analysis}\label{sec:error-anal}

\begin{table*}[h]
    \begin{adjustbox}{width=0.8\linewidth,center}
        \setlength{\tabcolsep}{6pt}
        \setlength{\cmidrulekern}{0.25em}
        \begin{tabular}{cccccccccc}
        \toprule
        \multirow{2}{4.5em}{\centering Error Type} & \multirow{2}{3em}{\centering Config} & \multicolumn{4}{c}{\gptf} & \multicolumn{4}{c}{\llama}\\\cmidrule(lr){3-6}\cmidrule(lr){7-10}
        & & \USEng & \IndEng & \Multi & \Trans & \USEng & \IndEng & \Multi & \Trans\\\midrule[\heavyrulewidth]
        \multirow{2}{4.5em}{\centering AD}%
            & PT & 18 (5) & 13 (3) & 13 (6) & 14 (6) & 10 (6) & 16 (10) & 18 (15) & 11 (7)\\
            & FT & \na (\na) & \na (\na) & \na (\na) & \na (\na) & 7 (6) & 9 (4) & 13 (13) & 11 (6)\\\midrule[\heavyrulewidth]
        \multirow{2}{4.5em}{\centering WD}%
            & PT & 4 (2) & 4 (4) & \na (\na) & 3 (3) & 3 (2) & 5 (5) & \na (\na) & 3 (3)\\
            & FT & \na (\na) & \na (\na) & \na (\na) & \na (\na) & 2 (2) & 5 (4) & \na (\na) & 3 (2)\\\midrule[\heavyrulewidth]
        \multirow{2}{4.5em}{\centering BDD}%
            & PT & 3 (2) & 16 (5) & \na (\na) & 7 (3) & \na (\na) & 2 (1) & \na (\na) & 3 (2)\\
            & FT & \na (\na) & \na (\na) & \na (\na) & \na (\na) & \na (\na) & 0 (0) & \na (\na) & 2 (3)\\\midrule[\heavyrulewidth]
        \multirow{2}{4.5em}{\centering CC}%
            & PT & 6 (2) & 5 (2) & 12 (5) & 4 (2) & 4 (2) & 5 (4) & 5 (3) & 2 (1)\\
            & FT & \na (\na) & \na (\na) & \na (\na) & \na (\na) & 2 (1) & 4 (2) & 3 (2) & 2 (0)\\\midrule[\heavyrulewidth]
        \multirow{2}{4.5em}{\centering PF}%
            & PT & 6 (2) & 2 (0) & 7 (4) & 4 (0) & 14 (8) & 3 (1) & 9 (5) & 4 (1)\\
            & FT & \na (\na) & \na (\na) & \na (\na) & \na (\na) & 7 (5) & 1 (1) & 5 (4) & 3 (0)\\\midrule[\heavyrulewidth]
        \multirow{2}{4.5em}{\centering ERR}%
            & PT & \na (\na) & \na (\na) & 6 (1) & \na (\na) & \na (\na) & \na (\na) & 4 (3) & \na (\na)\\
            & FT & \na (\na) & \na (\na) & \na (\na) & \na (\na) & \na (\na) & \na (\na) & 3 (1) & \na (\na)\\\midrule[\heavyrulewidth]
        \multirow{2}{4.5em}{\centering $\sum$}%
            & PT & 37 (13) & 40 (14) & 38 (16) & 32 (14) & 31 (18) & 31 (21) & 40 (29) & 23 (14)\\
            & FT & \na (\na) & \na (\na) & \na (\na) & \na (\na) & 18 (13) & 19 (11) & 27 (21) & 21 (11)\\
        \bottomrule
        \end{tabular}
    \end{adjustbox}
    \caption{Count of errors of \gptf and \llama for each subset. PT/FT: Pre-trained/Fine-tuned. `X (Y)' indicates that there are X errors in TWP and Y errors in TWS. $\sum$ is the sum of errors tagged in the sampled erroneous conversations by a model on a subset across all error types.}
    \label{tab:err}
\end{table*}

From \textbf{Test} set of each conversation subset, we randomly select 30 conversations that are mislabeled by \gptf and \llama, and manually analyse errors among all model variants across all subsets of conversations. We summarise the six error categories\footnote{Additonal examples for each error category are in Appendix~\ref{sec:errors}.} in Table~\ref{tab:err}. The error types are:

\paragraph{Ambigous Descriptions (AD)} This error type is observed when descriptions lack specificity (given the \textit{situational} constraint on the describer), leading to multiple potential answers. For the example target word--`\textit{engine},' the description provided is--`\textit{What we find in our. cars. in the front part?}'. Although these descriptions provide enough information to guide a human guesser to the right answer, they are often too vague to guide the LLM to a singular, correct interpretation.

\paragraph{Wrong Descriptions (WD)} 
These errors occur when the guesser guesses the target word even before the describer can finish the description completely. In the case of the target word `\textit{surname},' the model infers `\textit{parent}' when the description provided is--`\textit{beside your. uh. what is your elder? Uh what is}'. While human guessers might use their cognitive bias to guess correctly without the complete description, LLMs lack the ability to understand the target word from such a description.

\paragraph{Broken down description of prompt word (BDD)} This error occurs when the describer breaks down the target word into subwords and attempts to explain each separately. Generally, such descriptions involve longer turns. The guesser is then expected to piece together these fragments to deduce the original word, as in the case of the target word `\textit{Billie Holiday},' the describer individually describes the subwords `\textit{Billie}' and `\textit{Holiday}'. In such cases, LLMs sometimes latch onto the descriptions pertaining to later subwords, predicting a partially correct target word.

\paragraph{Shared Cultural Context (CC)} These errors arise when the human players use culturally shared notions in a conversation, often due to the describer's lack of familiarity with the target word. For example, an Indian describer explains the word `\textit{idli}' using examples of breakfast items and then asks the guesser to infer `\textit{Adele}'. The model is unable to understand this happening in the conversation.  

\paragraph{Public Figure (PF)} 
These errors pertain to inaccurate predictions generated by the model when the descriptions are about a well-known public figure. For example, the describer describes the target word '\textit{Mike Tyson}' as `\textit{Big guy that punched people out and he had a little bit of a lisp},' but the model generates `\textit{darth}'. 

\paragraph{Fallback Error (ERR)} While efforts were made to classify every mislabeled conversation into an error category, few generated target words were found to be inexact or inaccurate, even with apt descriptions in the conversations. For example, the target word--`\textit{Rose}' and the description--`\textit{This are the types of that's often given valentine day plant.}', the model generates `\textit{Gift}'. This example description mentions the word \textit{plant} which should have guided the model to a more specific target word than \textit{Gift}.

The error types \textbf{AD}, \textbf{CC}, and \textbf{PF} test the model's ability to predict the target word based on descriptions influenced by the describer's dialect, shared notions with the guesser, and perceived notions about the target word. Also, some of the conversations fall into multiple error categories except in the case of conversations in \textbf{ERR} (which is a mutually exclusive label).

Table \ref{tab:err} presents the error cases in `X (Y)' which indicates that there are X errors in TWP and Y errors in TWS for the corresponding configuration. The benefit of TWS providing options for the target word is seen in \textbf{AD}, where the alleviation is almost uniform across all dialects. The presence of direct or indirect references to the prompt word helps the LLM towards a plausible answer, in turn making it easier for them to choose an option. However, this error reduction does not extend to \textbf{CC}, which LLMs are unable to detect.

Fine-tuning helps to reduce the errors of \textbf{AD} category more for conversations of \IndEng dialect compared to \USEng. However, after removing the dialectal information, the conversations are insensitive to fine-tuning for the \textbf{AD} error cases. Additionally, fine-tuning helps to decrease errors in the \textbf{PF} category. As expected, it does not significantly reduce errors in the \textbf{WD} category.
\section{Related Work}\label{sec:related}
 Research in \textbf{dialect robustness} stems from the need for language technologies to be equitable and not reinforce any negative sentiments against a specific linguistic subgroup~\cite{blodgett-etal-2020-language}. LLMs perform poorly on several downstream tasks (such as the tasks in the GLUE benchmark) involving dialects other than mainstream US English~\cite{joshi2024natural, faisal2024dialectbench}. 
 
 Similar to our work, the evaluation of language understanding ability of LLMs has been explored using typical \textbf{conversation understanding tasks}~\cite{chen-etal-2022-unidu} like conversation summarisation~\cite{gliwa-etal-2019-samsum, chen-etal-2021-dialogsum}, conversation completion~\cite{sai-etal-2020-improving, ueyama-kano-2023-dialogue}, or NLU tasks~\cite{faisal2024dialectbench}. Other approaches involve conversation-based question-answering tasks that also evaluate the reasoning abilities of LLMs~\cite{sun-etal-2019-dream, qin-etal-2021-timedial}. Tasks like mask-filling were used to evaluate LLM-generated responses, more specifically~\citet{dey-desarkar-2023-dial} do so by making RoBERTa predict masked keyword utterances when given a context of dialogue history along with conditions like persona, topic, and facts. Different from standard language understanding tasks, ~\citet{chalamalasetti-etal-2023-clembench} presents a novel method to evaluate the ability of LLMs to act as `\textit{situational}' language understanding agents~\cite{schlangen-2023-general}. They do so by assigning roles to LLMs and generate dialogues resembling word games such as taboo, and test the language generating and instruction following abilities of LLMs based on the quality of game-play leading to successful target word prediction. 
 
 Although we propose a similar approach to evaluation by utilising conversations of such a word game, our work differs from theirs in two ways: (a) they use LLM-generated conversations while we rely on an existing dataset of conversations; (b) they do not employ dialects in their conversations while the dataset we use contains information about the dialects of the human speakers.
\section{Conclusion}
\label{sec:conclusion}
Although superlative performances have been reported on LLMs in recent times, recent work shows the performance gap between US English and other dialects of English. Our paper presents a first-of-its-kind evaluation of the multilingual LLMs for their robustness to minority language varieties, using their ability to predict target words in game-playing conversations. We use a dataset of target-word-masked conversations between US English speakers and those between Indian English speakers playing a game of taboo. We evaluate pre-trained and fine-tuned versions of one open-source and two closed-source models, on two tasks: target word prediction (TWP) and target word selection (TWS). Our results show that the LLMs indeed perform better for \USEng as compared to \IndEng on both tasks, with the average performance being higher by 12.66 and 17.4 on similarity and accuracy scores across all configurations. This shows that the LLMs, although multilingual, marginalise or discriminate against speakers of the Indian dialect. We also observe that pre-trained models report a degraded performance on conversations created using both rule-based (\Multi) and LLM-based (\Trans) transformations, as compared to their source conversations (\USEng and \IndEng respectively).  However, fine-tuning on \Multi yields a greater improvement in the task performances, as compared to that on \Trans.
This shows that the transformations that introduce dialectal information about a national variety help in improving the dialect robustness of LLMs more than the transformations that remove the said dialectal information.
Finally, our error analysis demonstrates that, while most errors are mitigated by providing options for masked target words (TWS; in both pre-trained and fine-tuned variants), multilingual LLMs struggle to interpret target words based on the shared cultural context between speakers.

Our extension \mmd is a dataset for TWP and TWS based on \md, consisting of four subsets: \USEng, \IndEng, \Multi, and \Trans. The dataset opens opportunities for future evaluations of dialect robustness using similar conversation-based tasks. Our evaluation methodology can also be scaled up and applied to other existing dialogue and discourse datasets, to evaluate the ability of LLMs on properties other than dialect robustness. 
\section*{Limitations}
The original \md paper states that their dataset may be dominated by Western entities to some degree. Therefore, it is possible that Indian speakers faced difficulties with the terms. Having said that, the instances selected for our dataset are the ones where the Indian players guessed the word correctly. We have not performed a detailed qualitative analysis of these conversations, except for a cursory sanity check. We also assume that the dialect of English from each locale is homogeneous. Assuming that \IndEng is the English spoken in every region of India is an unrealistic generalization of the diversity of dialects of English. 
In terms of model fine-tuning, our paper also does not cover the impact of quantization and different fine-tuning (including cross-dialect) techniques on the task.

\section*{Ethics Statement}
We use a publicly available dataset of conversations consisting of human players engaged in a game of taboo. The topics discussed in the dataset are fairly general and are unlikely to cause distress. The error analysis was performed by one of the authors of the paper. The AI-transformed (\Trans) conversations may contain biased output, arising due to inherent properties of GPT-based models. 
\bibliography{custom}
\appendix
\onecolumn
\section{Dataset Construction}\label{sec:masking}
Table~\ref{tab:mask} describes the example conversations from extended \md and their corresponding masked versions from \mmd. We mask the turn where the guesser utters the target word to help with formulating our downstream tasks. We mask the target word by finding the exact match in the conversation as shown in the conversations from Table~\ref{tab:mask}. In case of conversations where an exact match is not found (such as \textit{planets}), we find the utterance that is most similar to the target word using the similarity score\footnote{Described in Section~\ref{sec:metrics} of the main paper.}. The rest of the conversation is then pruned to make the masked target word (represented by `[MASK]') the last token in the conversation.


\begin{table*}[h!]
    \begin{adjustbox}{width=\linewidth,center}
        \renewcommand{\arraystretch}{1}
        \begin{tabular}{cp{18em}p{18em}}
            \toprule
            \multicolumn{1}{>{\centering\arraybackslash}m{4em}}{\textbf{Target Word}} & \multicolumn{1}{>{\centering\arraybackslash}m{18em}}{\textbf{\IndEng}} & \multicolumn{1}{>{\centering\arraybackslash}m{18em}}{\textbf{Masked \IndEng}} \\\midrule[\heavyrulewidth]
            \multirow{4}{4em}{\centering Fisherman} & \texttt{{Describer:} Uh. What do you call if we, what will be there in the water?} & \texttt{{Describer:} Uh. What do you call if we, what will be there in the water?}\\
            & \texttt{{Guesser:} Fishes} & \texttt{{Guesser:} Fishes}\\
            & \texttt{{Describer:} Who will catch that?} & \texttt{{Describer:} Who will catch that?}\\
            & \texttt{{Guesser:} \textit{\underline{\textbf{Fisherman}}}.} & \texttt{{Guesser:} \textit{\underline{\textbf{[MASK]}}}}\\\midrule[\heavyrulewidth]
            \multicolumn{1}{>{\centering\arraybackslash}m{4em}}{\textbf{Target Word}} & \multicolumn{1}{>{\centering\arraybackslash}m{18em}}{\textbf{\USEng}} & \multicolumn{1}{>{\centering\arraybackslash}m{18em}}{\textbf{Masked \USEng}} \\\midrule[\heavyrulewidth]
            \multirow{3}{4em}{\centering Planet} & \texttt{{Describer:} These are hard words. um Okay. So there's. the Sun and the Moon and all the rest of them.} & \texttt{{Describer:} These are hard words. um Okay. So there's. the Sun and the Moon and all the rest of them.}\\
            & \texttt{Guesser: And all the \textit{\underline{\textbf{planet}}}s?} & \texttt{Guesser: \textit{\underline{\textbf{[MASK]}}}}\\
            & \texttt{\textbf{(}Describer: Yes.\textbf{)}} & \\
            \bottomrule
        \end{tabular}
    \end{adjustbox}
    \caption{Masking conversations from the extended \md to create \mmd. The text such as \texttt{\textit{\textbf{\underline{this}}}} represents the target word utterance by the guesser which is masked (represented by, \texttt{\textit{\textbf{\underline{[MASK]}}}} in the \mmd version of the conversation. The rest of the original conversation is pruned as represented text in parentheses.}
    \label{tab:mask}
\end{table*}
\section{Transformation Issues}\label{sec:issues}
We present examples of transformation issues faced while creating \Trans in Table~\ref{tab:trans}. We create \Trans by prompting\footnote{The exact prompt can be found in Section~\ref{sec:method} of the main paper.} \gptf to remove exaggerations and dialectal information from \IndEng conversations. Table~\ref{tab:mult} presents examples of similar issues faced while creating \Multi using Multi-VALUE. As mentioned in Tables~\ref{tab:trans} and \ref{tab:mult}, a \textit{`typical'} transformed conversation maintains the semantic meaning but only differs from the original conversation grammatically. A \textit{`bad'} example deviates largely from the expected output. An \textit{`erroneous'} example is a result of Multi-VALUE not being able to transform a conversation from \USEng. Both \textit{`bad'} and \textit{`erroneous'} examples are excluded from the final set of conversations used in our evaluation.
\begin{table*}[h!] 
    \begin{adjustbox}{width=\linewidth,center}
        \renewcommand{\arraystretch}{1}
        \begin{tabular}{cp{18em}p{18em}}
            \toprule
            \multicolumn{1}{>{\centering\arraybackslash}m{6em}}{\textbf{Type}} & \multicolumn{1}{>{\centering\arraybackslash}m{18em}}{\textbf{\IndEng}} & \multicolumn{1}{>{\centering\arraybackslash}m{18em}}{\textbf{\Trans}} \\\midrule[\heavyrulewidth]
            \multirow{4}{6em}{\centering Typical}
                & \texttt{\textbf{Describer:} \textbf{(}Uh\textbf{)}. What do you call \textit{\underline{\textbf{if we, what will be there}}} in the water?} & \texttt{\textbf{Describer:} \textbf{(}$\varnothing$\textbf{)} What do you call \textit{\underline{\textbf{the creatures}}} in the water?}\\
                & \texttt{\textbf{Guesser:} Fish\textbf{(}es\textbf{)}} & \texttt{\textbf{Guesser:} Fish\textbf{(}$\varnothing$\textbf{)}.}\\
                & \texttt{\textbf{Describer:} Who \textit{\underline{\textbf{will catch that}}}?} & \texttt{\textbf{Describer:} Who \textit{\underline{\textbf{catches them}}}?}\\
                & \texttt{\textbf{Guesser:} Fisherm\textit{\underline{\textbf{a}}}n.} & \texttt{\textbf{Guesser:} Fisherm\textit{\underline{\textbf{e}}}n.}\\\midrule[\heavyrulewidth]
            
            \multirow{10}{6em}{\centering Bad}
                & \texttt{\textbf{Describer:} There. is a. there is a character in a movie} & \multirow{10}{18em}{The character being described is from a well-known movie and is known for the line ``I am still gorgeous.'' This character is similar to those found in Marvel movies and has a very muscular physique, which is achieved through scientific injections that cause his body to grow. The character being guessed is Captain America.}\\
                & \texttt{\textbf{Guesser:} um} & \\
                & \texttt{\textbf{Describer:} It's a very famous movie and it's a very. where is a where you can see famous dialogue called I am still gorgeous} & \\
                & \texttt{\textbf{Guesser:} uh. ok. uh} & \\
                & \texttt{\textbf{Describer:} character name. compare like Marvel movie} & \\
                & \texttt{\textbf{Guesser:} So. uh} & \\
                & \texttt{\textbf{Describer:} very muscular body gives scientifically. injections} &\\
                & \texttt{\textbf{Guesser:} ok} & \\
                & \texttt{\textbf{Describer:} His body will grow} & \\
                & \texttt{\textbf{Guesser:} uh Captain America} & \\
            \bottomrule
        \end{tabular}
    \end{adjustbox}
    \caption{Example \textit{transformations} of \IndEng to \Trans. We utilise \gptf to generate the response. The text in parentheses refers to the omission/removal of certain filler and exaggerated words, and the text such as \texttt{\textit{\textbf{\underline{this}}}}, refers to the words or sentences that were rephrased to convey the original meaning.}
    \label{tab:trans}
\end{table*}

\begin{table*}[h!] 
    \begin{adjustbox}{width=\linewidth,center}
        \renewcommand{\arraystretch}{1}
        \begin{tabular}{cp{18em}p{20em}}
            \toprule
            \multicolumn{1}{>{\centering\arraybackslash}m{6em}}{\textbf{Type}} & \multicolumn{1}{>{\centering\arraybackslash}m{18em}}{\textbf{\USEng}} & \multicolumn{1}{>{\centering\arraybackslash}m{18em}}{\textbf{\Multi}} \\\midrule[\heavyrulewidth]
            \multirow{2}{6em}{\centering Typical} 
                & \texttt{\textbf{Describer:} Perfect. Oh! \textbf{(}We\textbf{)} earn this. We go to our jobs.} & \texttt{\textbf{Describer:} Perfect. Oh! \textbf{(}$\varnothing$\textbf{)} \textbf{\underline{[are]}} earn\textbf{\underline{[ing]}} this. We \textbf{\underline{[are]}} go\textbf{\underline{[ing]}} to our jobs.}\\
                & \texttt{\textbf{Guesser:} Money} & \texttt{\textbf{Guesser:} Money}\\\midrule[\heavyrulewidth]
            
            \multirow{4}{6em}{\centering Error}
                & \texttt{\textbf{Describer:} This person. is in. oh films. It's a man. He's um. famous for a fine show in the '80s.} & \multirow{4}{18em}{\centering None}\\
                & \texttt{\textbf{Guesser:} Um. what else is he in?} & \\
                & \texttt{\textbf{Describer:} He's in a lot of things. I think he's kind of short. Some people think that he looks nice but I don't. Mhm. Mish Mission Impossible. Mission Impossible.} & \\
                & \texttt{\textbf{Guesser:} Tom Cruise} & \\
            \bottomrule
        \end{tabular}
    \end{adjustbox}
    \caption{Example \textit{transformations} of \USEng to \Multi. We utilise \gptf to generate the response. The text in parentheses refers to the omission/removal of words, and the text such as \texttt{\textbf{\underline{[this]}}}, refers to the dialectal features added using Multi-VALUE.}
    \label{tab:mult}
\end{table*}
\section{Errors}\label{sec:errors}
Table~\ref{tab:errors} describes additional examples for all identified error types\footnote{Defined in Section~\ref{sec:error-anal} of the main paper.}. As mentioned, each conversation can be classified under multiple error types. For example, the conversation about the target word--\textit{`Ryan Reynolds'} is classified as \textbf{CC}, but can also be classified as \textbf{PF}.

\begin{table*}[h!] 
    \begin{adjustbox}{width=1\linewidth,center}
        \renewcommand{\arraystretch}{0.9}
        \begin{tabular}{ccp{35em}c}
            \toprule
            \multicolumn{1}{>{\centering\arraybackslash}m{2em}}{\textbf{Type}} & \multicolumn{1}{>{\centering\arraybackslash}m{6em}}{\textbf{Target Word}} & \multicolumn{1}{>{\centering\arraybackslash}m{18em}}{\textbf{Conversation}} & \multicolumn{1}{>{\centering\arraybackslash}m{6em}}{\textbf{Model Prediction}}\\\midrule[\heavyrulewidth]
            \multirow{4}{2em}{\centering \textbf{AD}}& \multirow{2}{6em}{\centering Fisherman} & \texttt{{Describer:} Okay. Okay. A. guy um wants to um okay. Guy catching something in the water.} & \multirow{2}{2em}{\centering Fish}\\
            & & \texttt{{Guesser:} [MASK]} & \\\cmidrule(lr){2-4}
            & \multirow{2}{6em}{\centering Mike Tyson} & \texttt{Describer: Big guy that punched people out and he had a little bit of a lisp.} & \multirow{2}{6em}{\centering Darth Wader}\\
            & & \texttt{Guesser: [MASK]} & \\\midrule[\heavyrulewidth]
            \multirow{8}{2em}{\centering \textbf{WD}}& \multirow{6}{6em}{\centering Atlantic Ocean} & \texttt{Describer: One of the. of the of world.} & \multirow{6}{6em}{\centering Kanyakumari}\\
            & & \texttt{Guesser: Of the seventh wonder of the world. Taj mahal? Is it regarding sea?} & \\
            & & \texttt{Describer: No no no the. Towards the bottom of India.} & \\
            & & \texttt{Guesser: Is it regarding} & \\
            & & \texttt{Describer: what we have?} & \\
            & & \texttt{Guesser: [MASK]} & \\\cmidrule(lr){2-4}
            & \multirow{2}{6em}{\centering Beg} & \texttt{Describer: Um so if you don't have any money uh you may stand on the corner.} & \multirow{2}{6em}{\centering Panhandle}\\
            & & \texttt{Guesser: [MASK]} & \\\midrule[\heavyrulewidth]
            \multirow{14}{2em}{\centering \textbf{BDD}}& \multirow{10}{6em}{\centering Russian Language}& \texttt{Describer: Ok. Ah Largest continent in the world} & \multirow{10}{6em}{\centering Russian}\\
            & &  \texttt{Guesser: Ok.} & \\
            & &  \texttt{Describer: Ah like area wise. Which country?} & \\
            & &  \texttt{Guesser: Largest. vast area. vast area? Russia but.} & \\
            & &  \texttt{Describer: We need to add N over there at the end.} & \\
            & &  \texttt{Guesser: Russian} & \\
            & &  \texttt{Describer: We speak} & \\
            & &  \texttt{Guesser: What they speak?} & \\
            & &  \texttt{Describer: Yeah. Ok.} & \\
            & &  \texttt{Guesser: [MASK]} & \\\cmidrule(lr){2-4}
            & \multirow{4}{6em}{\centering Cold War} & \texttt{Describer: This is a two-word term. The first word is a common illness that causes a runny nose.} & \multirow{4}{6em}{\centering War}\\
            & & \texttt{Guesser: Cold.} & \\
            & & \texttt{Describer: Yes that's the first word. The second word refers to a conflict between two countries.} & \\
            & & \texttt{Guesser: [MASK]} & \\\midrule[\heavyrulewidth]
            \multirow{12}{2em}{\centering \textbf{CC}}& \multirow{10}{6em}{\centering Ryan Reynolds} & \texttt{Describer: It is like. One of the. Pen name. which we used in school school days.} & \multirow{10}{6em}{\centering Flair}\\
            & & \texttt{Guesser: Cello point pen. Fine Grip} & \\
            & & \texttt{Describer: No no no} & \\
            & & \texttt{Guesser: Reynolds} & \\
            & & \texttt{Describer: Uh yeah yeah} & \\
            & & \texttt{Guesser: This is a second word or first word.} & \\
            & & \texttt{Describer: Yeah this is second word} & \\
            & & \texttt{Guesser: First word is. Name} & \\
            & & \texttt{Describer:Yeah name related to the same} & \\
            & & \texttt{Guesser: [MASK]} & \\\cmidrule(lr){2-4}
            & \multirow{2}{6em}{\centering Mark Wahlberg} & \texttt{Describer: Okay. Um. He was the original. of the Funky Bunch. But then he stopped music.} & \multirow{2}{6em}{\centering Marky} \\
            & & \texttt{Guesser: [MASK]} & \\\midrule[\heavyrulewidth]
            \multirow{10}{2em}{\centering \textbf{PF}}& \multirow{6}{6em}{\centering Steve Jobs} & \texttt{Describer: Ok. He is a famous person and he is a. a. for. what we call? Um now it is a. Its. giving competition to Android. what we call?} & \multirow{6}{6em}{\centering Steve}\\
            & & \texttt{Guesser: ok. so he is the fond ok sorry } & \\
            & & \texttt{Describer: he is a founder of so and so company. Its a U. S. company} & \\
            & & \texttt{Guesser: so it is giving competition to Android means Google ok.. So} & \\
            & & \texttt{Describer: and he is the founder of that company} & \\
            & & \texttt{Guesser: [MASK]} & \\\cmidrule(lr){2-4}
            & \multirow{4}{6em}{\centering Kanye West} & \texttt{Describer: All right.} & \multirow{4}{6em}{\centering Clint} \\
            & & \texttt{Guesser: How do you wanna skip that one} & \\
            & & \texttt{Describer: All right. Now um. This guy he um. He just bought a ranch in Wyoming.} & \\
            & & \texttt{Guesser: [MASK]} & \\\midrule[\heavyrulewidth]
            \multirow{2}{2em}{\centering \textbf{ERR}}& \multirow{2}{6em}{\centering Podium} &  \texttt{Describer: Okay um. uh. well I isn't sure I'm not sure but uh letting are seeing. Well it's like preacher are churching. I am standing behind this. uh. in in used for speaker.} & \multirow{2}{6em}{\centering Pulpit}\\
            & &  \texttt{Guesser: [MASK]} & \\
            \bottomrule
        \end{tabular}
    \end{adjustbox}
    \caption{Example conversations (\textit{`Conversation'}) for each error type (\textit{`Type'}) along with the reference target word (\textit{`Target Word'}) and the generated target word (\textit{`Model Prediction'}). All model predictions are generated using the pre-trained variants of \gptf and \llama.}
    \label{tab:errors}
\end{table*}

\end{document}